\documentclass[a4paper, 10pt, conference]{IEEEtran} 

\IEEEoverridecommandlockouts
\usepackage{cite}
\usepackage{amsmath,amssymb,amsfonts}
\usepackage{algorithmic}
\usepackage{graphicx}
\usepackage{textcomp}
\usepackage{xcolor}

\usepackage{graphicx}
\usepackage{booktabs}
\usepackage{multirow}
\usepackage{hyperref}
\usepackage{subcaption}
\usepackage{cleveref}

\crefname{equation}{Eq.}{Eqs.}     
\crefname{figure}{Fig.}{Figs.}     
\crefname{table}{Tab.}{Tabs.}

\def\BibTeX{{\rm B\kern-.05em{\sc i\kern-.025em b}\kern-.08em
    T\kern-.1667em\lower.7ex\hbox{E}\kern-.125emX}}
\begin{document}

\title{Evaluation of Polarimetric Fusion for Semantic Segmentation in Aquatic Environments
\thanks{This research was partially supported by the French Agence Nationale de la Recherche, under grant ANR-23-CE23-0030 (project R3AMA)}
}

\author{
\IEEEauthorblockN{Luis F. W. Batista\textsuperscript{*}, Tom Bourbon\textsuperscript{*}, C{\'e}dric Pradalier}
\IEEEauthorblockA{\textit{Georgia Tech-Europe - IRL2958 GT-CNRS, Metz, France} \\
luis.batista@gatech.edu, tbourbon@georgiatech-metz.fr}
\thanks{\textsuperscript{*}These authors contributed equally to this work.}
}

\maketitle

\begin{abstract}
Accurate segmentation of floating debris on water is often compromised by surface glare and changing outdoor illumination. Polarimetric imaging offers a single-sensor route to mitigate water-surface glare that disrupts semantic segmentation of floating objects.
We benchmark state-of-the-art fusion networks on PoTATO, a public dataset of polarimetric images of plastic bottles in inland waterways, and compare their performance with single-image baselines using traditional models.
Our results indicate that polarimetric cues help recover low-contrast objects and suppress reflection-induced false positives, raising mean IoU and lowering contour error relative to RGB inputs. These sharper masks come at a cost: the additional channels enlarge the models increasing the computational load and introducing the risk of new false positives.
By providing a reproducible, diagnostic benchmark and publicly available {code}\footnote{https://github.com/luisfelipewb/EvalPolFusion}, we hope to help researchers choose if polarized cameras are suitable for their applications and to accelerate related research.

\end{abstract}


\section*{Introduction}

Perception is fundamental to autonomous systems, enabling effective interpretation and interaction with their environments. Semantic segmentation, in particular, plays a critical role by providing pixel-level classification, essential for precise localization and object identification in complex scenes\cite{minaee2022segmentation}.
However, achieving robust semantic segmentation in outdoor, water-rich environments remains a significant challenge due to water surface reflections and dynamic lighting conditions. These conditions obscure object boundaries and degrade the performance of standard vision algorithms.

Individual sensors often exhibit limitations under specific environmental conditions, and therefore, multimodal approaches have emerged to address these weaknesses by integrating data from different sensor types, thereby enhancing robustness and reliability \cite{nawaz2024fusionsurvey}. Various strategies for combining different modalities involve trade-offs in terms of computational complexity, synchronization requirements, and the ability to capture cross-modal correlations.

There has been growing interest in polarimetric imaging. Microgrid-based polarimetric sensors capture polarization and color data synchronously, avoiding the alignment issues common to multi-sensor systems\cite{wang2022microgridtech}. This makes them particularly suitable for outdoor applications, where abundant natural polarization sources, such as skylight scattering and water surface reflections, produce rich polarization patterns \cite{foster2018polarisation}.

\begin{figure}[t!]
\centering
\begin{subfigure}{0.325\linewidth}
    \includegraphics[width=\linewidth]{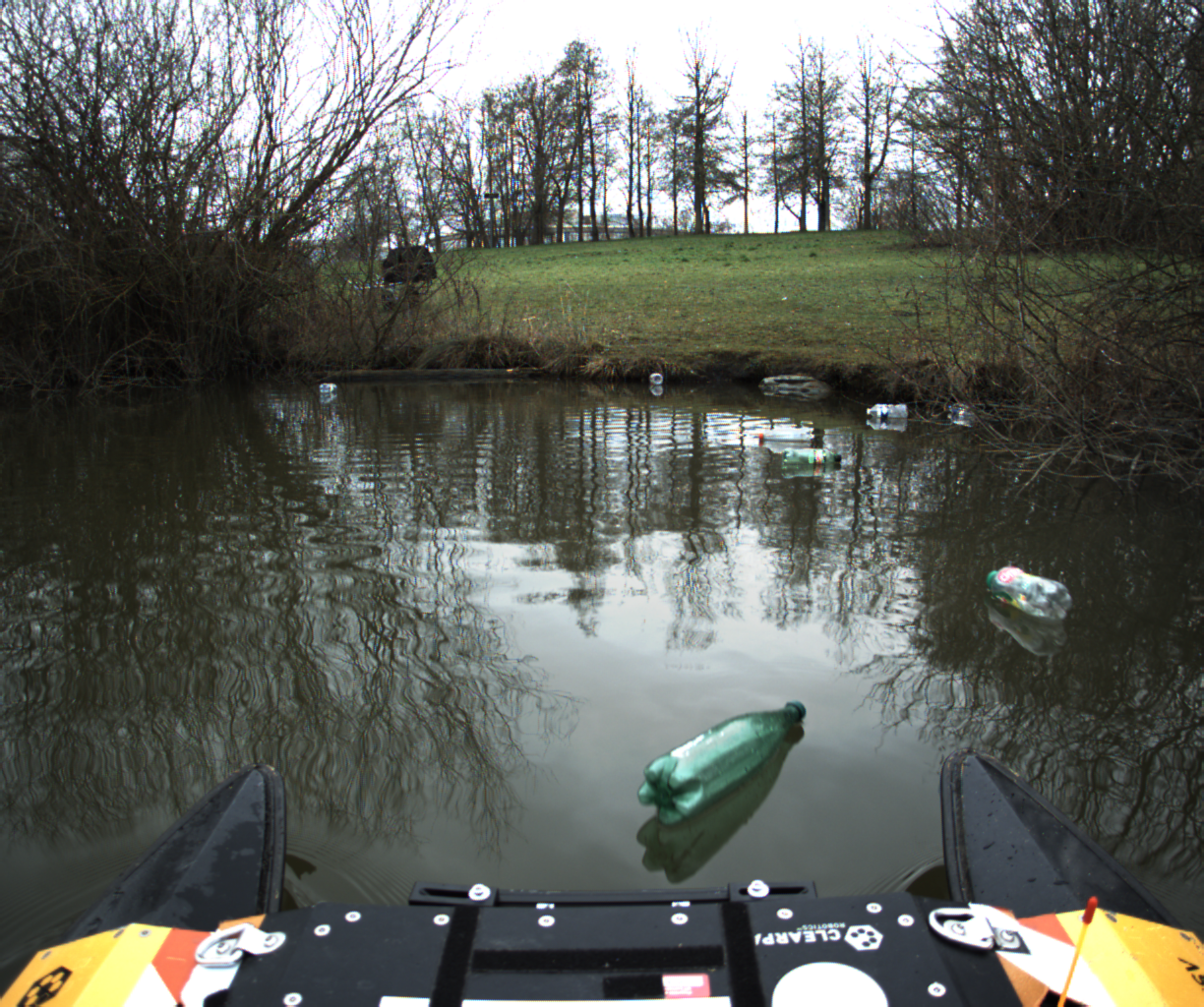}
    \caption{\textit{RGB}.}
    \label{fig:rgb_example}
\end{subfigure}
\hfill
\begin{subfigure}{0.325\linewidth}
    \includegraphics[width=\linewidth]{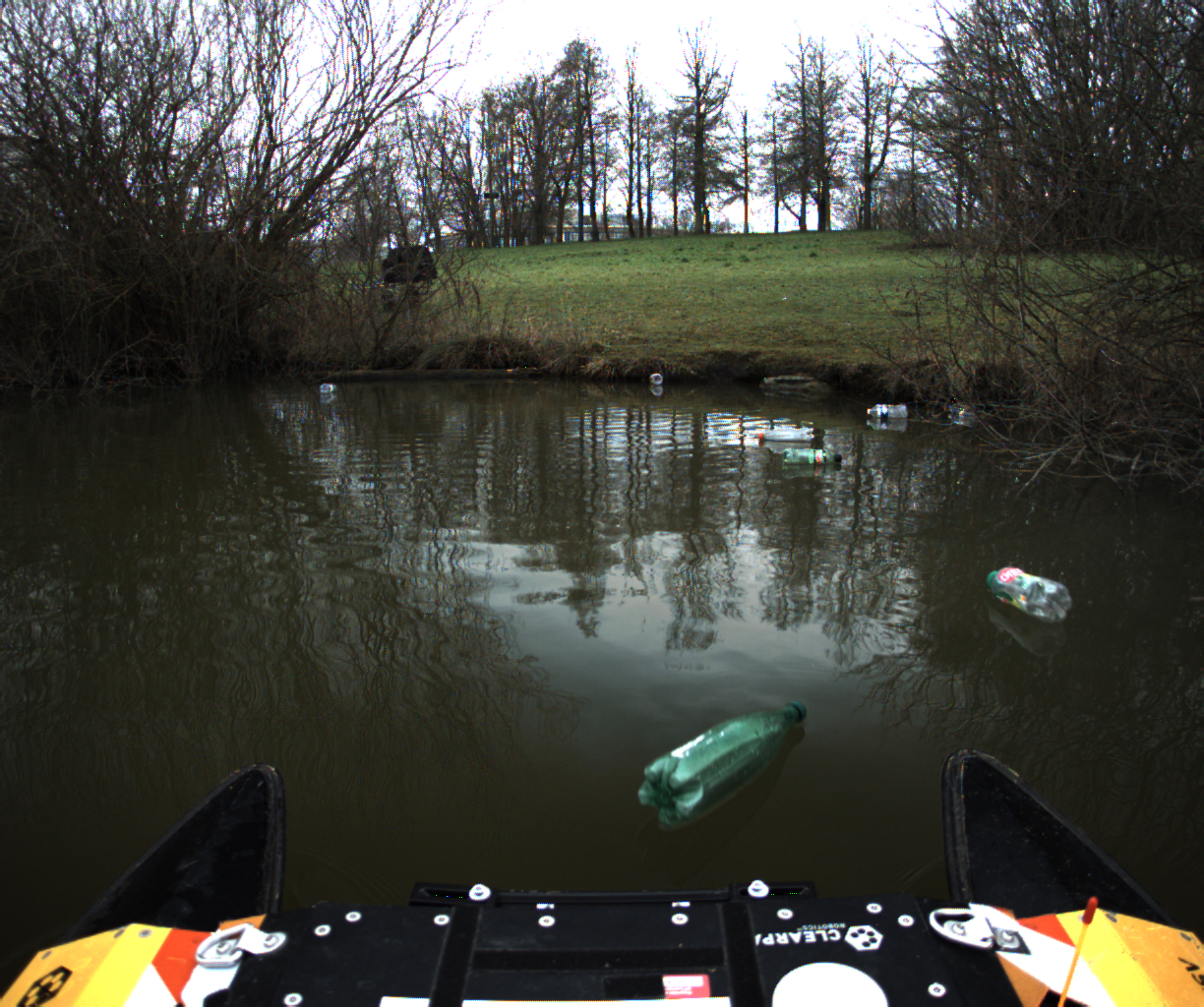}
    \caption{\textit{DIF}.}
    \label{fig:dif_example}
\end{subfigure}
\hfill
\begin{subfigure}{0.325\linewidth}
    \includegraphics[width=\linewidth]{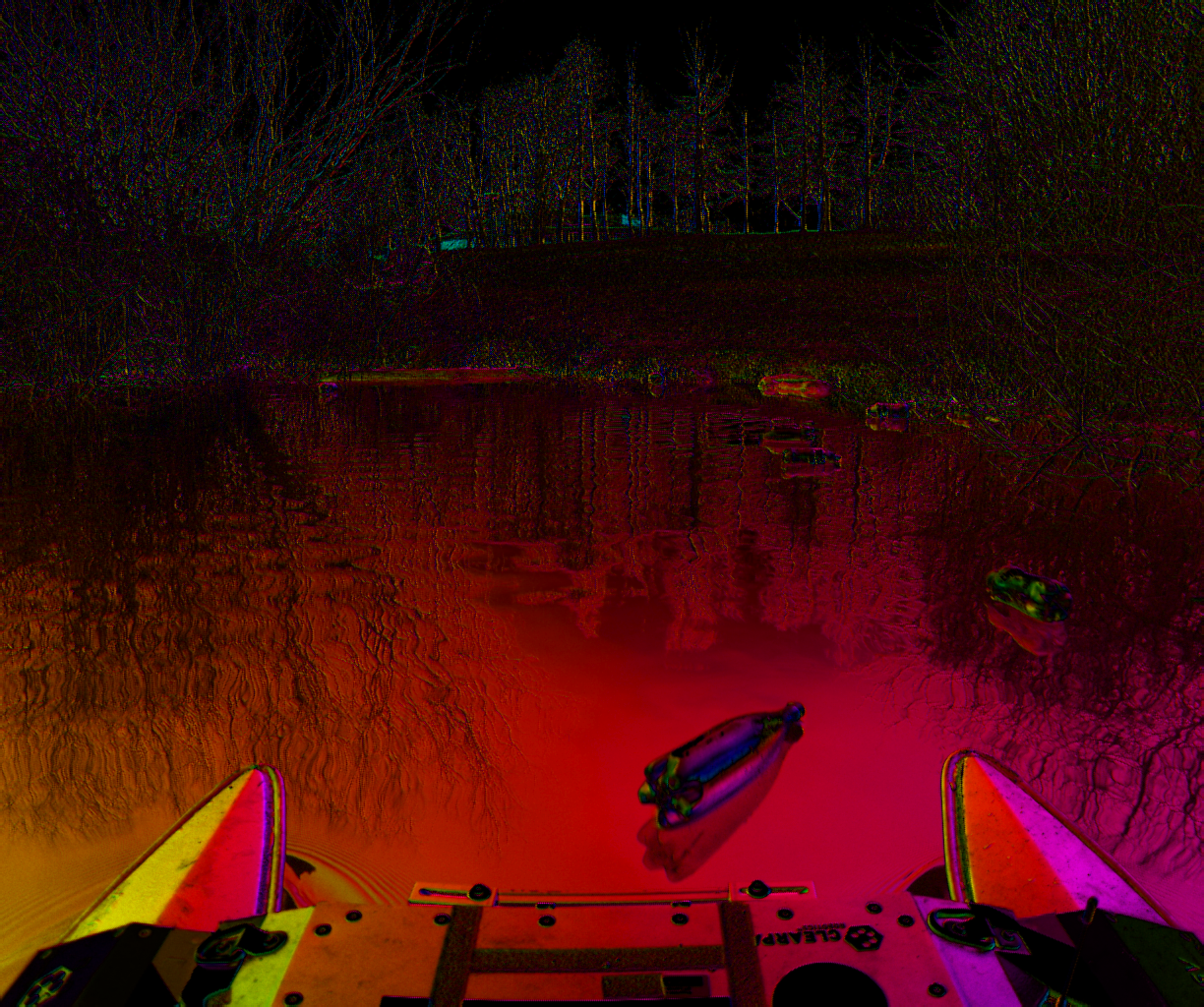}
    \caption{\textit{POL}.}
    \label{fig:pol_example}
\end{subfigure}
\caption{Samples from the PoTATO dataset. \Cref{fig:rgb_example} Standard color image. (\Cref{fig:dif_example}) Diffuse-only image with surface glare suppressed. (\Cref{fig:pol_example}) Pseudo-color image encoding (\textit{AoLP}) and (\textit{DoLP) information}}
\label{fig:example}
\vspace{-10pt}
\end{figure}

In this work, we investigate the potential of polarimetric imaging for semantic segmentation of floating objects in water-rich environments. Specifically, we focus on evaluating fusion algorithms that integrate chromatic and polarimetric information to enhance perception. Our main contributions are summarized as follows:

\begin{itemize}
\item Establish baseline for well-known semantic segmentation architectures using polarimetric-aware images.
\item Conduct a comprehensive evaluation of state-of-the-art fusion algorithms for combining color and polarimetric modalities, specifically Angle of Linear Polarization (\textit{AoLP}) and Degree of Linear Polarization (\textit{DoLP}).
\item Analyze the practical potential and limitations of applying these methods in autonomous systems, with particular attention to performance gains and computational overhead.
\end{itemize}

\section*{Related Work}

According to \cite{zhang2021surveymarineobjectdetection}, detecting floating objects is often hindered by changing outdoor lighting, sun glare, and reflections on the water surface. A comprehensive survey of maritime computer vision datasets and deep learning techniques is provided in \cite{trinh2025visionsurvey}. The authors emphasize that unpredictable reflections on the water surface remain a key bottleneck, arguing for further research to improve the stability of deep learning models in response to such fluctuating aquatic conditions.

To address these perception challenges, multi-sensor configurations have gained increasing attention in recent years. While the number of annotated datasets for semantic segmentation and object detection has expanded, smaller yet increasing collections of multi-modal sensor data have also begun to attract attention \cite{trinh2025visionsurvey}. Notable efforts include pairing mmWave radar with RGB cameras for floating-bottle detection \cite{cheng2021flow}, and fusing RGB, grayscale, and thermal cameras with both LiDAR and radar for maritime perception \cite{douguet2023mmodforboats}. Focusing on inland waterways, Cheng \textit{et al.} \cite{cheng2021usvmultisensordataset} introduce a benchmark comprising multiple radars, a stereo camera, and a LiDAR. However, the temporal and spatial misalignment typical of heterogeneous sensors increases fusion complexity and can degrade overall performance.

As an alternative to multi-sensor setups, polarization imaging has gained growing attention due to its ability to capture complementary information using a single, synchronous sensor. A recent survey highlights the rapid expansion of polarimetric imaging research within deep learning \cite{li2023polarimetric}. Polarization signals have also improved the segmentation of transparent objects \cite{kalra2020deep}, yet these experiments were performed on indoor images that do not capture challenging outdoor lighting, and their implementation is not publicly released.

In contrast, recent architectures offer more flexible multi-modal fusion frameworks and are often accompanied by publicly available code and datasets. MCubeS \cite{liang2022multimodal} provides outdoor RGB and polarimetric imagery but focuses on material recognition, with its water class limited to small urban puddles. CMX \cite{zhang2023crossmodalfusion} augments an RGB backbone by concatenating a single additional modality channel. Extending this idea, CMNeXt \cite{zhang2023cmnext} reports substantial gains by flexibly fusing RGB with arbitrary modalities, thus accommodating more than two inputs. MMSFormer \cite{reza2024mmsformer} and StitchFusion \cite{li2024stitchfusion}, can also integrate an arbitrary number of modalities and achieve improved results when evaluated on the MCubeS dataset. Although encouraging, these studies investigate polarization fusion mainly for autonomous-driving or outdoor scenes in limited datasets that contain little or no water surfaces.

Despite the versatility of recent fusion architectures, polarization has rarely been explored in water-rich environments. Recent work on polarimetric object detection based on the PoTATO dataset \cite{batista2024potato} claims that leveraging \textit{DoLP} (D) and \textit{AoLP} (A) information improves robustness to specular reflections. While the results indicate the value of polarimetric signals in aquatic scenes, the PoTATO benchmark currently evaluates each modality independently, leaving the potential benefits of fusion unexplored.

Our work addresses the lack of polarimetric fusion studies in aquatic environments. Using the publicly available PoTATO dataset, we establish single-modality segmentation baselines and benchmark state-of-the-art fusion algorithms.

\section*{Methodology}

\subsection{Dataset Preparation}

We base our experiments on the PoTATO dataset \cite{batista2024potato}, which contains raw polarimetric images of plastic bottles floating on water. This section details the derivation of polarimetric modalities from the raw data, the conversion of bounding-box annotations to pixel-accurate segmentation masks, and the construction of the final training, validation and test partitions.

The raw images in the PoTATO dataset contain the intensity of light at four polarization orientations, $I_{0}$, $I_{45}$, $I_{90}$ and $I_{135}$, for each pixel of the RGGB Bayer pattern. These measurements allow the computation of the Stokes vector (\Cref{eq:stokes}). 
\begin{equation}
\label{eq:stokes} 
\begin{bmatrix} 
S_0 \\ S_1 \\ S_2
\end{bmatrix}
=
\begin{bmatrix}
I_{0}+I_{90}\\ I_{0} - I_{90}\\ I_{45} - I_{135}
\end{bmatrix}
\end{equation}

From the Stokes components we derive three scalar quantities that leverage polarimetric information: the Degree of Linear Polarization, $\mathrm{DoLP} = \sqrt{S_1^{2}+S_2^{2}}/S_0$; the Angle of Linear Polarization, $\mathrm{AoLP} = \tfrac12\arctan(S_2/S_1)$; and an approximation of the diffuse-only intensity, $I_{\text{dif}} = (S_0-\sqrt{S_1^{2}+S_2^{2}})/2$. Following the extraction pipeline of \cite{batista2024potato}, these signals are rendered as three images: RGB (standard color), DIF (reflection-suppressed using $I_{\text{dif}}$), and POL (pseudo-color composite with \textit{AoLP} as hue and \textit{DoLP} as value).

Because PoTATO provides bounding-box annotations only, we produced pixel-level masks with the Segment Anything Model (SAM) \cite{kirillov2023sam} prompted with the original boxes. All masks were inspected and, where required, adjusted manually. Reflections on the water surface were excluded whenever distinguishable from the bottle.

Distant bottles appear as small objects, occupying only a few pixels in the image. To mitigate this scale imbalance, we cropped each image around the region of interest, removing irrelevant sky, vegetation, and boat hull regions. The resulting images have a resolution of $1224 \times 512$ pixels. We refer to this final version of the dataset as PoTATO-Seg, which comprises 1,500 images divided into 1,000 for training, 200 for validation, and 300 for testing.

\subsection{Segmentation Architectures}

\textbf{Fusion models}. We evaluate CMNeXt \cite{zhang2023crossmodalfusion}, MMSFormer \cite{reza2024mmsformer} and StitchFusion \cite{li2024stitchfusion}. Each network is trained with an RGB image, with different polarization modalities resulting in four input sets: RGB, RGBD, RGBA and RGBAD.
\textbf{Single-image baselines}. To benchmark fusion, we fine-tuned U-Net \cite{ronneberger2015unet} and DeepLabV3 \cite{chen2019rethinking} on each image type (RGB, DIF and POL). Supplying a single image to any fusion backbone simplifies the network to a SegFormer-style encoder-decoder \cite{xie2021segformer}; therefore, a dedicated SegFormer baseline was not included. Together these three models cover convolutional, atrous-convolutional and transformer paradigms, providing a comprehensive reference for the fusion experiments.

\subsection{Experimental Details}
Training was carried out on two NVIDIA RTX 3090 with 24 GB of memory each. To preserve the physical consistency of the polarimetric information, we applied only horizontal flipping as data augmentation. Because many ground-truth masks are small, no additional down-sampling was performed; instead, each cropped region was resized to $512 \times 512$ before being passed to the network.
Each model was optimised for 50 epochs with a batch size of 4. The learning rate was linearly warmed up during the first 10 epochs and then scaled by a factor of 0.01. Full configuration files and scripts are available in the public code repository.

\section*{Results}

In this section, we present our quantitative and qualitative analysis. All metrics and examples are from the test set.

Single-image baseline metrics in \Cref{tab:baseline_metrics} show that input modality influences segmentation quality for every architecture. DIF images yield the highest mIoU and F1 scores for all models except DeepLab, where the polarimetric input (POL) is slightly superior. Additionally, the modern fusion-style networks StitchFusion, MMSFormer, and CMNeXt surpass the classical U-Net and DeepLab. These findings confirm that suppressing water-surface glare through diffuse-only preprocessing provides consistent benefit, and polarimetric cues are a valuable complementary source of information for segmentation tasks. Our baselines reproduce literature results \cite{batista2024potato}, and validate the experimental setup.

\begin{table}[thb]
    \caption{Single-image baselines on the PoTATO-Seg test set.}
    \centering

\begin{tabular}{llrrrr}
\toprule
Model & Modality & mIoU(\%) & Precision & Recall & F1 \\
\midrule
U-Net & RGB & 73.034 & 0.841 & 0.839 & 0.833 \\
      & POL & 71.518 & 0.887 & 0.770 & 0.812 \\
      & DIF & \textbf{75.786} & 0.875 & 0.848 & 0.854 \\
\midrule
DeepLab & RGB & 70.946 & 0.838 & 0.818 & 0.817 \\
        & POL & \textbf{71.752} & 0.874 & 0.783 & 0.819 \\
        & DIF & 70.922 & 0.851 & 0.796 & 0.814 \\
\midrule
CMNeXt & RGB & 76.081 & 0.868 & 0.861 & 0.857 \\
       & POL & 73.394 & 0.832 & 0.848 & 0.832 \\
       & DIF & \textbf{76.646} & 0.884 & 0.850 & 0.859 \\
\midrule
MMSFormer & RGB & 75.679 & 0.867 & 0.857 & 0.853 \\
          & POL & 72.361 & 0.819 & 0.849 & 0.825 \\
          & DIF & \textbf{76.729} & 0.882 & 0.853 & 0.860 \\
\midrule
StitchFusion & RGB & 77.043 & 0.871 & 0.866 & 0.862 \\
             & POL & 71.944 & 0.833 & 0.830 & 0.821 \\
             & DIF & \textbf{77.197} & 0.875 & 0.865 & 0.863 \\
\bottomrule
\end{tabular}
\label{tab:baseline_metrics}
\end{table}

The metrics from fusion models shown in \Cref{tab:fusion_metrics} indicate that adding \textit{DoLP} and \textit{AoLP} channels to RGB images generally enhances segmentation, although the magnitude of the gain depends on the fusion approach. StitchFusion benefits most: each auxiliary channel increases mIoU, and the RGBDA variant achieves the highest overall scores. MMSFormer sees moderate improvements that plateau when both cues are present, whereas CMNeXt shows only minor, non-significant gains. 
These findings suggest that architectures incorporating explicit cross-modal fusion modules exploit multi-channel inputs more effectively. Moreover, polarimetric cues provide complementary information, although the benefits of their integration vary across models.

\begin{table}[thb]
    \caption{Fusion models including polarimetric cues.}
    \centering

\begin{tabular}{llrrrr}
\toprule
Model & Modality & mIoU(\%) & Precision & Recall & F1 \\
\midrule
StitchFusion & RGB & 77.04 & 0.871 & 0.866 & 0.862 \\
             & RGBD & 78.71 & 0.894 & 0.862 & 0.874 \\
             & RGBA & 78.86 & 0.900 & 0.860 & 0.875 \\
             & RGBDA & \textbf{79.03} & 0.902 & 0.862 & 0.876 \\
\midrule
MMSFormer & RGB & 75.68 & 0.867 & 0.857 & 0.853 \\
          & RGBD & 77.11 & 0.857 & 0.876 & 0.862 \\
          & RGBA & \textbf{78.20} & 0.896 & 0.856 & 0.871 \\
         & RGBDA & 77.42 & 0.883 & 0.855 & 0.864 \\
\midrule
CMNeXt & RGB & 76.08 & 0.868 & 0.861 & 0.857 \\
       & RGBD & 76.47 & 0.870 & 0.860 & 0.859 \\
       & RGBA & 76.43 & 0.888 & 0.843 & 0.857 \\
       & RGBDA & \textbf{76.74} & 0.883 & 0.849 & 0.860 \\
\bottomrule
\end{tabular}
\label{tab:fusion_metrics}
\end{table}

\begin{figure}[tbh]
\centering
    \includegraphics[width=\linewidth]{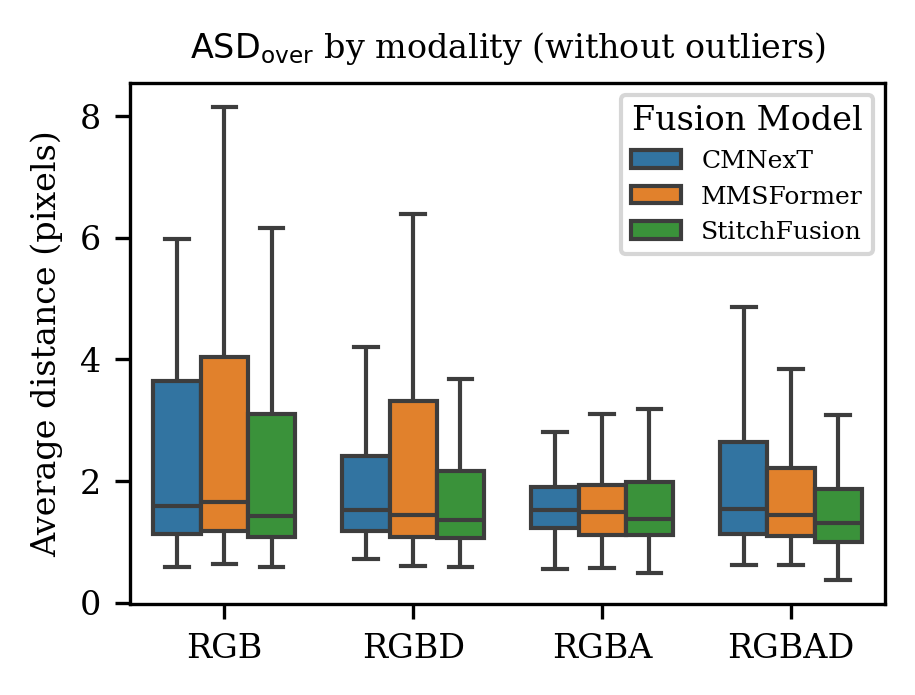}
    \caption{Adding polarimetric channels lowers boundary error.}
    \label{fig:metrics}
\end{figure}

To quantify boundary accuracy across modalities, we computed the Average Surface Distance ($\text{ASD}_{\text{over}}$) defined as the mean of the shortest distances from each point on the prediction contour to the ground-truth contour. Smaller values denote better alignment. \Cref{fig:metrics} shows these values after discarding statistical outliers located more than $1.5\times\text{IQR}$ beyond the first or third quartile to eliminates a few rare false detections that would otherwise distort the distributions. The plots reveal a consistent trend across all three models: adding polarimetric channels lowers the $\text{ASD}_{\text{over}}$ metric, demonstrating reduced average contour error compared with RGB alone.

The mean inference time across the test set is presented in \Cref{tab:inference_times}. Adding extra modalities consistently slowed inference. While DeepLab is marginally faster than U-Net, both are significantly faster when compared to any fusion model.

\begin{table}[ht]
\centering
\begin{tabular}{c|c|c|c|c}
\hline
Model      & RGB & RGBD & RGBA & RGBAD \\
\hline
U-Net       & 5.2 ms      & -             & -             & -              \\
DeepLab    & \textbf{5.0 ms}      & -             & -             & -              \\
\hline
CMNeXt     & 24.3 ms      & 37.7 ms       & 37.8 ms       & 39.6 ms        \\
MMSFormer  & \textbf{18.2 ms}      & \textbf{24.2 ms}       & \textbf{24.3 ms}       & \textbf{24.5 ms}        \\
StitchFusion & 22.1 ms    & 45.4 ms       & 45.7 ms       & 83.9 ms        \\
\hline
\end{tabular}
\caption{Mean Inference Times (ms) for the PoTATO-Seg.}
\label{tab:inference_times}
\end{table}

\begin{figure*}[t!]
\centering
\begin{subfigure}{0.95\linewidth}
    \includegraphics[width=\linewidth]{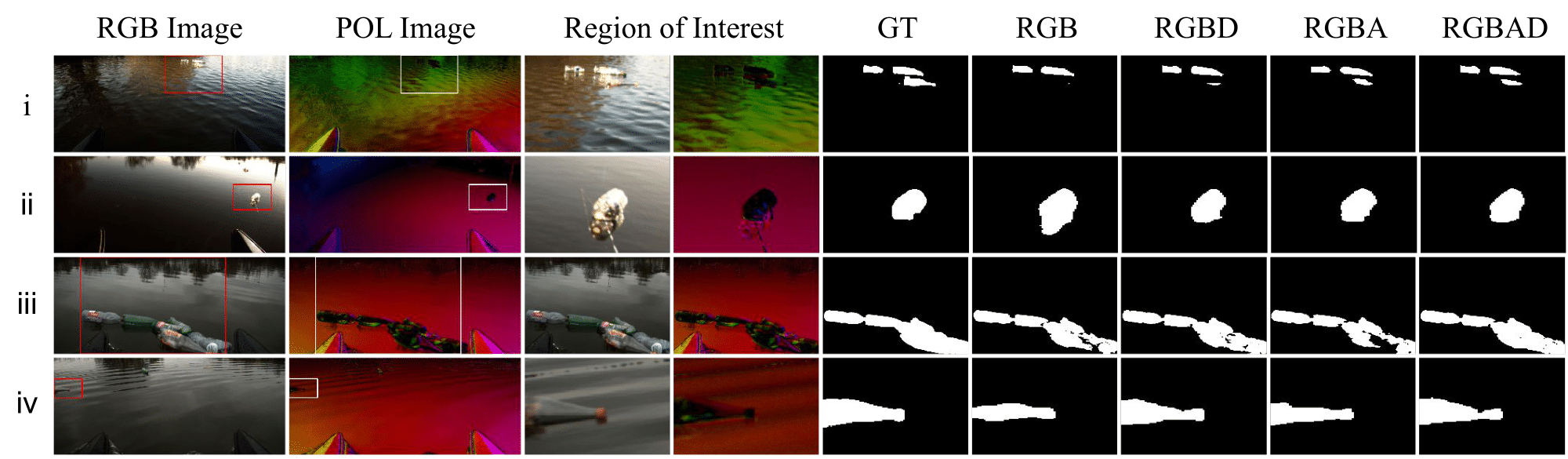}
    \caption{Examples when polarization enables detection of a bottle under low-contrast lighting (i), suppresses a strong reflection mistaken for an object in RGB (ii), and recovers a bottle whose color matches the water surface (iii, iv), all producing sharper masks for the RGBAD input.}
    \label{fig:qualitative-good}
\end{subfigure}
\begin{subfigure}{0.95\linewidth}
    \includegraphics[width=\linewidth]{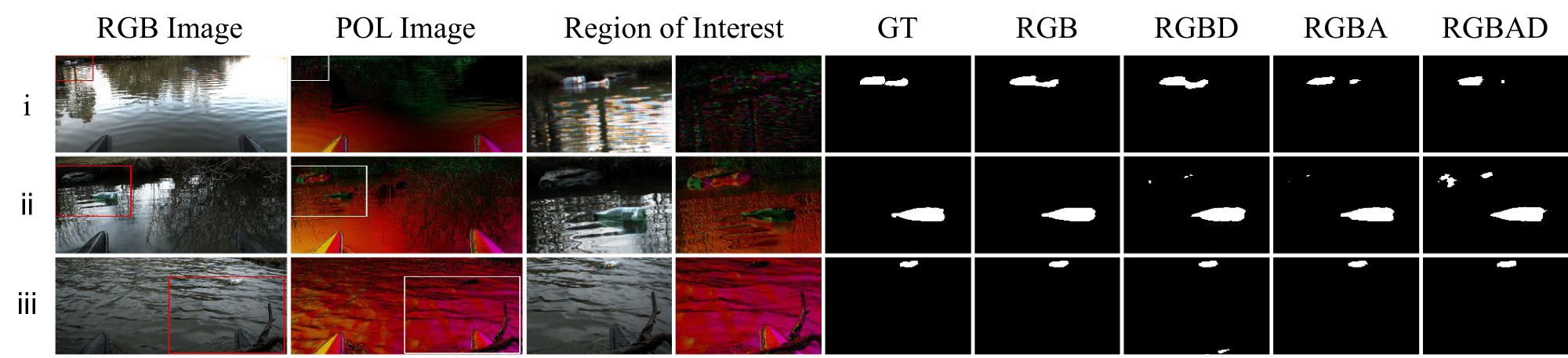}
    \caption{Polarimetric fusion limitations. (i) Failure to improve detection of a distant bottle located in a region with weak polarization signals. (ii) Shoreline rocks and (iii) tree branches produce polarimetric signatures similar to floating bottles, resulting in false positives.}
    \label{fig:qualitative-bad}
\end{subfigure}
    \caption{Qualitative Examples of Polarimetric Fusion using StitchFusion.}
    \label{fig:qualitative}
\end{figure*}

For qualitative assessment we examine StitchFusion, the best-performing model. \Cref{fig:qualitative} gathers scenes with the largest IoU gaps between modalities, complementing the quantitative results. \Cref{fig:qualitative-good} shows examples when polarization enables StitchFusion to detect low-contrast objects and suppress reflection-induced false positives. \Cref{fig:qualitative-bad} highlights failure cases in which the added channels introduce false positives or miss true positives in regions with weak polarization signals.

\section*{Discussion}

Baseline experiments demonstrate consistent results with what is presented in previous research \cite{batista2024potato} and fusion models gain most in regions with strong polarimetric information. \textit{DoLP} and \textit{AoLP} information can help to suppress glare and reveal color-concealed targets, with the largest improvements observed for StitchFusion, whose cross-modal blocks reduce contour error and increase IoU metrics. 

In some cases, extra channels can introduce false positives when background materials produce similar polarimetric responses and provide little benefit where polarization is weak. This highlights the opportunity for better fusion strategies. The lack of ImageNet-style pre-training forces the network to learn polarimetric features from scratch, slowing convergence. Additionally, the available polarimetric datasets are rather small. In our experiments, validation curves start to indicate overfitting after about 50 epochs, suggesting that the long schedules common in the fusion literature could be excessive for the dataset used in this evaluation.

Despite these training challenges, the contour analysis presented in \Cref{fig:metrics} indicates that polarimetric fusion yields high-precision masks. On the other hand, the extra channels enlarge the input tensor, increasing model size and latency. Inference time experiments confirm that these costs must be balanced against the accuracy gains, and this should be carefully considered, especially for applications with real-time requirements and limited computing power.

\section*{Conclusion}

In this paper, we evaluated how polarimetric cues can enhance semantic segmentation of floating objects. We provide a systematic and reproducible evaluation baseline that includes two standard segmentation models and three modern general-purpose fusion models.
Our experiments indicate that integrating \textit{DoLP} and \textit{AoLP} information raised mean IoU and lowered contour error by suppressing reflection and increasing contrast. Nonetheless, the fusion models sometimes produced false positives when background materials had similar polarimetric signatures and contributed little in areas with weak polarization.
We conclude that in general, polarimetric information is most useful for applications that demand precise masks, and have strong potential for application with rich polarimetric information, but they incur additional training complexity and increased inference time.
Future work should explore larger and more diverse polarimetric datasets, tailored training strategies, and lightweight fusion modules optimized for real-time deployment.

\bibliographystyle{IEEEtran}
\bibliography{IEEEabrv,references}

\vspace{12pt}

\end{document}